\documentclass[letterpaper, 10 pt, conference]{ieeeconf}
\IEEEoverridecommandlockouts
\usepackage{bbm}
\usepackage{siunitx}
\usepackage{cite}
\usepackage{bm}
\usepackage{color}
\usepackage{graphbox,graphicx}
\usepackage{stfloats}
\usepackage{amsmath}
\usepackage{amssymb}
\usepackage{calligra}
\usepackage{algorithmic}
\usepackage{array}
\usepackage{multirow}
\usepackage{tabularray}
\usepackage{makecell}
\usepackage{hhline}
\usepackage[utf8]{inputenc}
\usepackage[caption=false]{subfig}
\usepackage{fixltx2e}
\usepackage{kotex}
\usepackage{float}
\usepackage{eufrak}
\usepackage{supertabular}
\usepackage[table]{xcolor} 
\usepackage{tabularx}
\usepackage{soul}
\let\labelindent\relax
\usepackage{enumitem}
\usepackage{svg}
\usepackage{url}
\usepackage{booktabs}
\usepackage{threeparttable}
\usepackage[colorlinks=true, urlcolor=blue, linkcolor=red]{hyperref}
\usepackage{pifont}
\usepackage{adjustbox}
\usepackage{ulem}

\usepackage{nicematrix} 
\usepackage{tikz}  

\usepackage{textcase}

\usepackage{caption}
\title{\LARGE \bf
Beyond the Patch: Exploring Vulnerabilities of Visuomotor Policies \\ via Viewpoint-Consistent 3D Adversarial Object
}
\author{Chanmi Lee, Minsung Yoon, Woojae Kim, Sebin Lee, and Sung-eui Yoon$^{\dagger}$
\thanks{The authors are with the School of Computing at the Korea
Advanced Institute of Science and Technology (KAIST), Daejeon, 34141, Republic of Korea. 
E-mails: {\tt\scriptsize \{chanmi99, minsung.yoon, wkim97, seb.lee\}@kaist.ac.kr}, {\tt\scriptsize sungeui@kaist.edu}. 
$^{\dagger}$Corresponding author.}
}
 
\begin{document}

\maketitle

\thispagestyle{empty}
\pagestyle{empty}

\begin{abstract}
Neural network–based visuomotor policies enable robots to perform manipulation tasks but remain susceptible to perceptual attacks. For example, conventional 2D adversarial patches are effective under fixed-camera setups, where appearance is relatively consistent; however, their efficacy often diminishes under dynamic viewpoints from moving cameras, such as wrist-mounted setups, due to perspective distortions. To proactively investigate potential vulnerabilities beyond 2D patches, this work proposes a viewpoint-consistent adversarial texture optimization method for 3D objects through differentiable rendering. As optimization strategies, we employ Expectation over Transformation (EOT) with a Coarse-to-Fine (C2F) curriculum, exploiting distance-dependent frequency characteristics to induce textures effective across varying camera–object distances. We further integrate saliency-guided perturbations to redirect policy attention and design a targeted loss that persistently drives robots toward adversarial objects. Our comprehensive experiments show that the proposed method is effective under various environmental conditions, while confirming its black-box transferability and real-world applicability.

\end{abstract}

\section{Introduction} \label{sec:1} 
    Vision-based manipulation policies have gained significant attention for enabling robots to effectively interact with objects through visual understanding~\cite{zhu2022viola, rahmatizadeh2018vision}.     
    Among these, end-to-end approaches directly map visual inputs to actions, enabling robots to learn task-relevant features implicitly~\cite{chi2024universal, seo2025legato, zhao2023learning, fu2024mobile, lin2024data}. 
    However, their reliance on neural networks makes such policies inherently vulnerable to adversarial examples,
    carefully crafted inputs designed to induce unintended robot behaviors
    ~\cite{jia2022physical, chen2024diffusion}. 
    For instance, a malicious object unpacked from a shipment can deceive warehouse robots into hazardous actions, such as incorrect grasps or collisions.
    
\begin{figure}[!t] 
    \vspace{6pt}
    \centering
    \includegraphics[width=\linewidth]{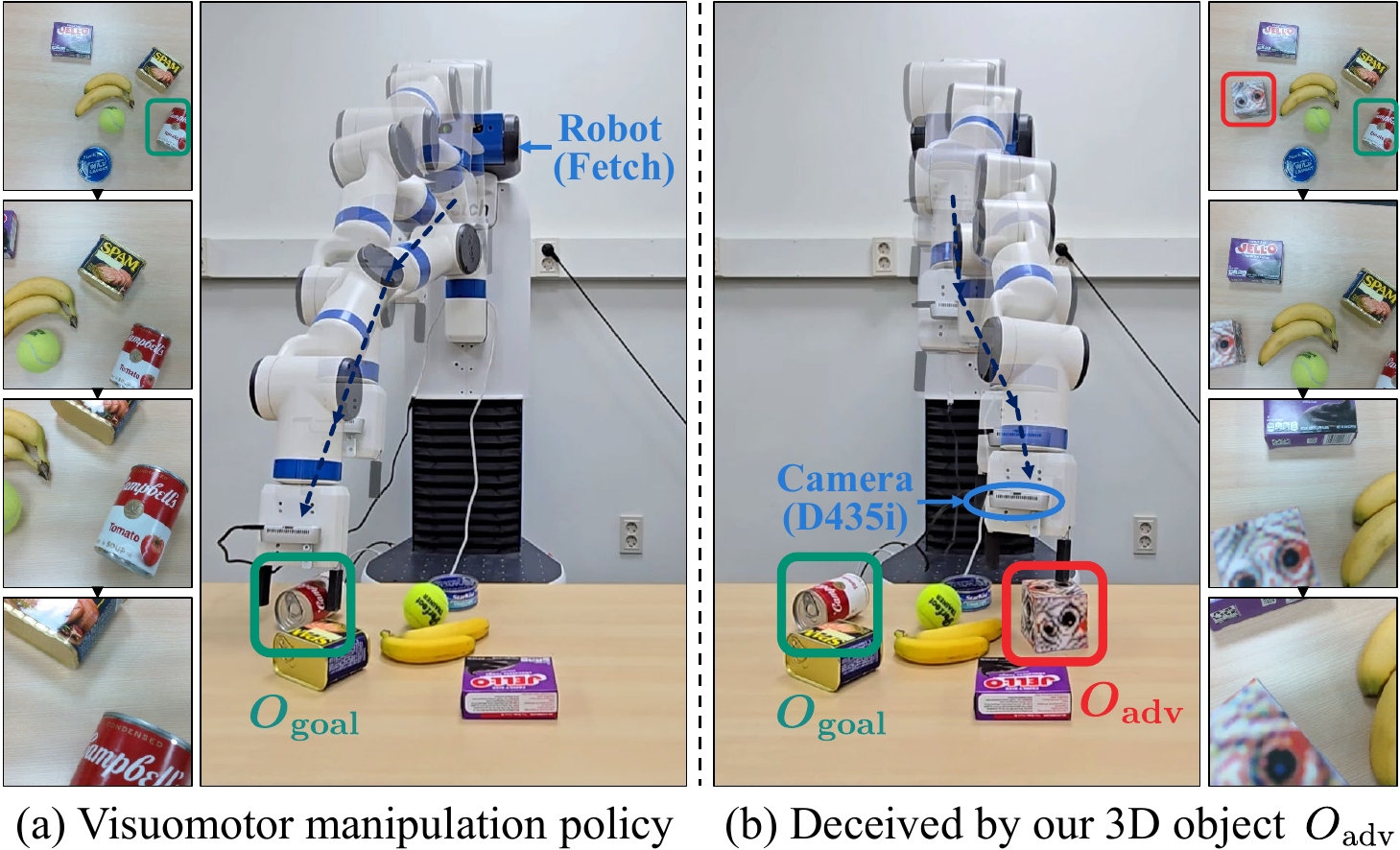} 
    \vspace{-13pt}
    \caption{Visuomotor policy deception using a 3D adversarial object. (a) The policy successfully guides the robot to its target $O_{\text{goal}}$. (b) Our adversarial object $O_{\text{adv}}$ manipulates the visual input, compelling the policy to misguide the robot towards itself instead of the true target $O_{\text{goal}}$.}
    \vspace{-17pt} 
    \label{fig:figure1} 
    \end{figure}

\begin{figure}[t!] 
    \centering
    \includegraphics[width=\linewidth]{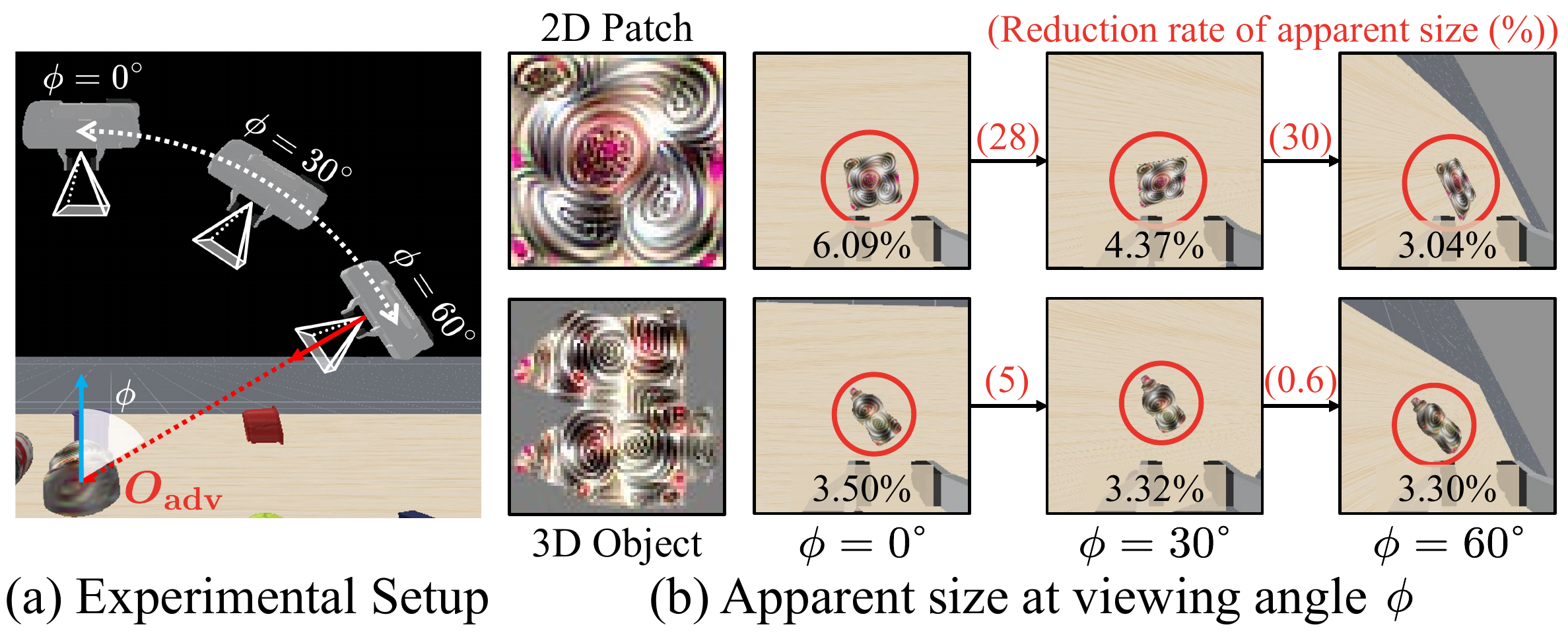} 
    \vspace{-13pt} 
    \caption{Apparent size comparison of a 3D adversarial object and a 2D patch. The 2D patch significantly shrinks and distorts, especially at large viewing angles $\phi$, unlike the more stable 3D object.}
    \vspace{-20pt} 
    \label{fig:3D2D} 
    \end{figure}

    Recent studies have begun exploring the vulnerability of visuomotor policies to adversarial attacks, primarily focusing on 2D adversarial patches~\cite{brown2017adversarial} due to their practical feasibility~\cite{jia2022physical, chen2024diffusion}. 
    While 2D patches have demonstrated adversarial efficacy primarily in confined settings with fixed third-person cameras, their performance diminishes in dynamic viewpoint scenarios involving wrist-mounted cameras or mobile platforms~\cite{chi2024universal, seo2025legato, zhao2023learning, fu2024mobile, lin2024data, li2024okami}. 
    These setups induce significant viewpoint shifts, which arise not only from variations in the initial robot poses but also from continuous robot movements (Fig.~\ref{fig:figure1}(a)).
    Under such 3D viewpoint variations, the inherent planarity of 2D patches cannot fully accommodate the 3D nature of the viewpoint shifts.
    The planar constraint causes appearance inconsistency and severe perspective distortion at oblique angles, neutralizing the adversarial pattern’s effectiveness (Fig.~\ref{fig:3D2D}).
    Thus, exploring viewpoint-consistent 3D adversarial objects becomes essential for evaluating security vulnerabilities and ensuring the reliability of visuomotor manipulation policies in real-world deployments.

    \textbf{Main Contributions.} 
    We propose a viewpoint-consistent 3D adversarial attack that effectively misleads visuomotor policies with wrist-mounted cameras.
    Our method optimizes a texture over a 3D mesh, leveraging Expectation over Transformation (EOT)~\cite{madry2018towards} to achieve consistent attack efficacy across varied initial robot poses and continuous robot movements (Fig.~\ref{fig:figure1}(b)).
    To achieve this, we design a targeted adversarial loss that persistently misguides the robot toward the object, ensuring the adversarial object remains in the camera view throughout the task.
    Moreover, the attack succeeds regardless of object pose, enabling versatile placement.

    To maximize attack potency, we introduce two key strategies: 
    (1) Coarse-to-Fine (C2F) Optimization: A hierarchical strategy that ensures consistent attack efficacy across varying camera distances.
    We first optimize global features from distant viewpoints, then progressively refine fine-grained details from closer perspectives. 
    (2) Saliency-Guided Attack: A strategy to redirect the policy's focus. 
    Using saliency maps to identify decision-critical regions, we optimize textures to shift attention from the true target to the adversarial object.
    Our extensive experiments demonstrate that our method not only maintains high attack efficacy under diverse viewpoint conditions but also transfers successfully to black-box scenarios and real-world applications. 
    To the best of our knowledge, this is the first systematic analysis of visuomotor manipulation policy vulnerability to 3D adversarial attacks.
    
\section{Related Works} \label{sec:sec2}

    \subsection{Adversarial Examples on Robotic Manipulation Systems}
    Adversarial examples induce model malfunctions by adding perturbations to neural network inputs~\cite{szegedy2013intriguing, goodfellow2014explaining}.
    In white-box attack scenarios where model architectures and parameters are fully accessible, gradient-based methods such as FGSM~\cite{goodfellow2014explaining} and PGD~\cite{madry2018towards} can efficiently compute optimal perturbations by leveraging the loss function's gradients to maximize prediction errors.
    To enhance real-world deployability, adversarial patches localize these perturbations into printable patterns that can be physically placed in environments, creating visible yet deceptive 2D patches~\cite{brown2017adversarial}.

    Recently, adversarial attacks have expanded beyond static computer vision tasks to dynamic robot systems,
    with a growing focus on the vulnerabilities of manipulation policies.
    Some approaches disrupt motion planners or Model Predictive Control (MPC) by physically manipulating the arrangement of objects in the robot's workspace~\cite{wu2024characterizing, agarwal2023synthesizing}.
    By strategically configuring the environment, these methods cause the robot to misinterpret spatial layouts, leading to path-planning errors or an obstructed field of view.

    In parallel, extensive research has focused on directly perturbing visual inputs to robotic policies, inducing unintended behaviors. Early works altered only one or a few pixels or applied perturbations across entire RGB images~\cite{alharthi2024physical, huang2023trade}. Furthermore, physically deployable 2D adversarial patches placed within the robot workspace demonstrated the susceptibility of object detection models and visuomotor policies to physical adversarial attacks~\cite{jia2022physical, chen2024diffusion}. In particular, Chen et al.~\cite{chen2024diffusion} applied random affine transformations to enhance the physical robustness of these patches; however, fundamentally, 2D patches struggle to consistently handle viewpoint changes inherent to 3D spaces. 
    Especially in robotic manipulation environments using wrist-mounted cameras, where viewpoint changes occur frequently due to robot arm movements~\cite{chi2024universal, seo2025legato, zhao2023learning, fu2024mobile, lin2024data}, the effectiveness of 2D-based attacks is significantly limited, making research into viewpoint-robust attack techniques essential.

    \subsection{3D Adversarial Examples}
    To develop adversarial attacks robust to diverse camera viewpoints, computer vision research has focused on various 3D adversarial attack methods optimizing textures or shapes of 3D objects~\cite{athalye2018synthesizing, zeng2019adversarial, wang2022fca, byun2022improving, suryanto2022dta,huang2024towards}. These methods typically utilize Expectation over Transformation (EOT)~\cite{athalye2018synthesizing}, differentiable renderers~\cite{zeng2019adversarial, wang2022fca, suryanto2022dta, byun2022improving}, or Neural Radiance Fields (NeRF)~\cite{huang2024towards} to ensure consistent performance across different perspectives.
    
    Recent research in autonomous driving has also actively explored adversarial attacks considering viewpoint variations. Studies leveraging NeRF have optimized 3D object textures to effectively evaluate vulnerabilities in 3D object detection models and vision-based driving policies under varying viewpoints~\cite{li2024adv3d, abeysirigoonawardena2023generating}. Additionally, Chahe et al.~\cite{chahe2023dynamic} highlighted the necessity of dynamic patches capable of maintaining consistent attack despite varying distances and angles from vehicles.
    
    Ensuring viewpoint robustness is equally critical in robotic manipulation environments where frequent viewpoint changes occur. However, unlike autonomous driving, manipulation tasks face unique challenges: (1) dynamically updating patches (\textit{e.g.}, digital screens) is impractical in typical manipulation setups, and (2) objects are often randomly scattered on surfaces like tabletops, causing viewpoint variations and significant variations in image appearance, requiring adversarial objects to maintain effectiveness regardless of their unpredictable placement. Considering these constraints, we propose optimizing 3D adversarial object textures to robustly analyze the vulnerabilities of visuomotor manipulation policies across diverse viewpoints and random object placements.

\section{Methodology}\label{sec:method}

    \subsection{Problem Formulation}
    \label{sec:prob_form}

    In this work, we attack an end-to-end visuomotor manipulation policy that employs a neural network $\pi_\omega$.
    This policy maps image observations $I$ from an eye-in-hand camera to actions $\mathbf{a}=\pi_\omega(I)$, where $\omega$ denotes the network parameters.
    The robot executes this policy to perform manipulation tasks (\textit{e.g.}, reaching a target object $O_\text{goal}$).

    Against such policies, we propose a viewpoint-consistent 3D adversarial attack under a white-box scenario.
    We aim to craft an adversarial mesh object $O_\text{adv}$ that consistently disrupts or prevents the robot from reaching the target object $O_\text{goal}$, regardless of the viewing angles encountered during task execution.
    To achieve this, we optimize a texture pattern for mapping onto the object's mesh surface.
    This adversarial texture is designed to misguide the robot's manipulation throughout the entire task execution by misleading the policy's visual perception.

    \subsection{Gradient-based Texture Optimization}
    \label{sec:gradient_opt} 
\begin{figure*}[t!] 
    \centering
    \includegraphics[width=1\textwidth, trim=0 1.2cm 0 0, clip]{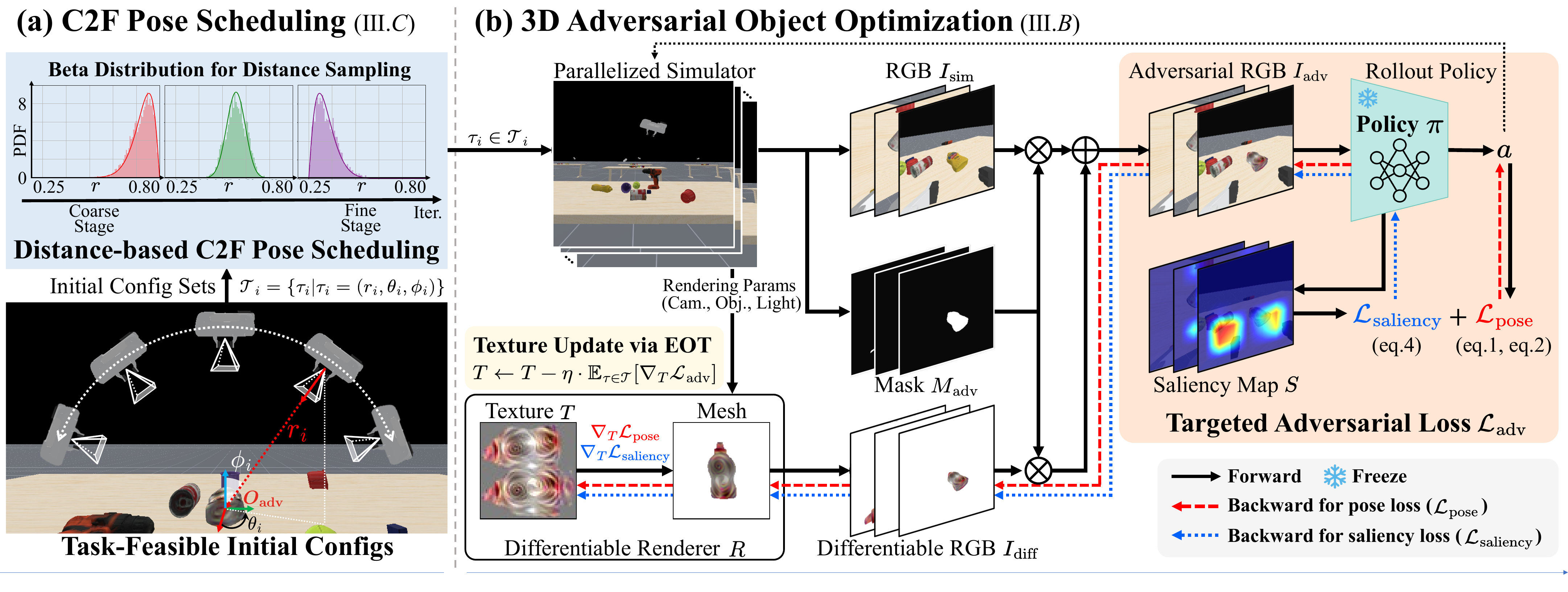} 
    \vspace{-12pt}
    \caption{Overview of the proposed method. (a) Coarse-to-Fine (C2F) Pose Scheduling: From a set of task-feasible initial configurations, poses where the original policy succeeds, we schedule viewpoint sampling using a distance-based Beta distribution. The scheduler progressively shifts focus from distant (Coarse stage) to closer (Fine stage) viewpoints.
    (b) 3D Adversarial Object Optimization Pipeline: Guided by the Expectation over Transformation (EOT) framework, the pipeline optimizes the adversarial texture $T$ through short policy rollouts from each initial pose $\tau_i$.  
    In each step, the policy $\pi_\omega$ processes a composite image $I_\text{adv}$ (formed from $I_\text{diff}$ and $I_\text{sim}$) to output an action. 
    A targeted adversarial loss is then computed from the resulting action, guiding the robot toward the adversarial object $O_\text{adv}$.
    The total loss, reflecting actual image-action pairs from the rollout, is backpropagated to update the texture $T$.
    }
    \vspace{-15pt} 
    \label{fig:figure2} 
    \end{figure*}

    In this section, we present a gradient-based optimization method for adversarial object texture that maintains consistent attack effectiveness throughout manipulator movements. 
    Our approach is driven by formulating an adversarial objective function that integrates pose alignment and model attention guidance, with optimization performed using Expectation over Transformation (EOT). 
    Fig.~\ref{fig:figure2} illustrates our overall attack framework, which consists of two main stages: Coarse-to-Fine (C2F) scheduling and the 3D adversarial attack pipeline.

    \noindent
    \subsubsection{\textbf{Targeted Pose Loss}}
    Unlike approaches that assume static cameras, our method addresses constantly changing viewpoints induced by the robot's own movement. 
    In such scenarios, the adversarial object $O_\text{adv}$ can fall outside the camera's field of view throughout the robot's movement.
    To maintain adversarial effectiveness throughout the entire trajectory, we design a targeted pose loss $\mathcal{L}_\text{pose}$ to consistently misguide the robot end-effector toward $O_\text{adv}$.

    The pose loss consists of two components: (1) an orientation loss $\mathcal{L}_\text{ori}$ and (2) a distance loss $\mathcal{L}_\text{dist}$.
    The orientation loss $\mathcal{L}_{\text{ori}}$ encourages the end-effector to point toward the adversarial object.
    This is achieved by maximizing the cosine similarity between the end-effector's intended heading vector $\mathbf{v}_\text{ee}$ that would result from action $\mathbf{a}$, and a target vector $\mathbf{v}_\text{target}$ pointing from the end-effector's action-intended position $\mathbf{p}_\text{next}$ toward $O_\text{adv}$:
    \begin{equation}
        \label{eq:ori_loss}
        \mathcal{L}_{\text{ori}} = 1 - \frac{\mathbf{v}_\text{ee} \cdot \mathbf{v}_\text{target}}{\| \mathbf{v}_\text{ee} \| \| \mathbf{v}_\text{target} \|} .
    \end{equation}
    The distance loss $\mathcal{L}_\text{dist}$ minimizes the Euclidean distance between the adversarial object position $\mathbf{p}_\text{adv}$ and end-effector action-intended position $\mathbf{p}_\text{next}$:
    \begin{equation}
        \label{eq:dist_loss}
        \mathcal{L}_{\text{dist}} = \|\mathbf{p}_\text{adv} - \mathbf{p}_\text{next}\|_2 .
    \end{equation}
    The total pose loss $\mathcal{L}_\text{pose}$ combines both terms via the weight $\lambda_\text{dist}$ such that $\mathcal{L}_\text{pose} = \mathcal{L}_\text{ori} + \lambda_\text{dist} \cdot \mathcal{L}_\text{dist}$.

    \subsubsection{\textbf{Targeted Saliency-guidance Loss}}
    In addition to the pose loss, we further redirect the policy's visual attention from the original goal $O_{\text{goal}}$ toward the adversarial object $O_{\text{adv}}$.
    We employ gradient-based saliency maps $S$, inspired by Grad-CAM~\cite{selvaraju2017grad}, which 
    highlight the image regions the policy focuses on to output an action.
    The saliency map is derived from the feature maps $A \in \mathbb{R}^{C \times H \times W}$ of the policy's visual backbone. We compute an importance weight $w_k$ for each channel by averaging gradients of the action norm $\|\mathbf{a}\|_2$ with respect to activations $A$:
    \begin{equation}
        w_k = \frac{1}{H \times W} \sum_{i,j} \frac{\partial \|\mathbf{a}\|_2}{\partial A_k(i,j)}; \, 
        S = \mathrm{ReLU} \left( \sum_k w_k A_k \right).
    \end{equation}
    The saliency loss is formulated to maximize the average saliency over the adversarial object region while minimizing it over the goal object region:
    \begin{equation}
        \mathcal{L}_\text{saliency} = - \frac{\sum_{i,j} (S \odot M_{\text{adv}})_{i,j}}{\sum_{i,j} (M_{\text{adv}})_{i,j}} + \frac{\sum_{i,j} (S \odot M_{\text{goal}})_{i,j}}{\sum_{i,j} (M_{\text{goal}})_{i,j}},
    \end{equation}
    where $M_{\text{adv}}$ and $M_{\text{goal}}$ are binary masks of $O_{\text{adv}}$ and $O_{\text{goal}}$, respectively.

    Thus, the final adversarial loss is defined as: 
    $\mathcal{L}_\text{adv} = \mathcal{L}_\text{pose} + \lambda_\text{saliency}\cdot\mathcal{L}_\text{saliency}$, where $\lambda_\text{saliency}$ balances the two terms. 
    To ensure that the gradients from $\mathcal{L}_\text{pose}$ and $\mathcal{L}_\text{saliency}$ do not conflict with each other during optimization, we utilize the Projecting Conflicting Gradients (PCGrad)~\cite{yu2020gradient} algorithm. PCGrad resolves potential conflicts by projecting the gradient of one loss onto the other and removing any opposing components before the update.

    \noindent
    \subsubsection{\textbf{Expectation over Transformation}}
    \label{sec:eot}
    To ensure our adversarial object $O_\text{adv}$ is effective from various viewpoints, we use the Expectation over Transformation (EOT)~\cite{madry2018towards} framework.   
    This involves optimizing the object's texture $T$ via gradient-based optimization, to minimize our adversarial loss $\mathcal{L}_{\text{adv}}$ over a diverse distribution of transformations $\mathcal{T}$:
    \begin{equation}
    T^* = \underset{T}{\operatorname{arg\,min}} \,\,\mathbb{E}_{\tau=(r,\theta,\phi) \sim \mathcal{T}}\left[\mathcal{L}_{\text{adv}}(T, \tau)\right],
    \end{equation}
    where $\tau=(r,\theta,\phi)$ denotes transformations defined by distance $r$, azimuth angle $\theta$, and polar angle $\phi$ between the adversarial object $O_\text{adv}$ and the robot end-effector poses.

    To ground the optimization in realism, we dynamically construct the transformation distribution $\mathcal{T}$ by performing short rollouts that capture the robot's actual behavior as it is influenced by the evolving adversarial texture $T_t$. 
    These rollouts originate from a set of initial configurations, where each configuration defines the initial poses for the adversarial object, the end-effector, and all other scene objects, including goals and obstacles.
    The scene configurations are parameterized as an initial relative pose $\tau_i=(r_i,\theta_i,\phi_i)$ between the end-effector and the adversarial object. 
    We select only configurations where the original policy successfully completes its task, focusing the optimization on meaningful scenarios.
    
    The expected gradient is approximated using the collected transformations $\mathcal{T}$:
    ${\mathbf{g}}_t \! =\! \mathbb{E}_{(r,\theta,\phi)\sim\mathcal{T}}\left[\nabla_{T_t}\mathcal{L}_{\text{adv}}(T_t, r, \theta, \phi)\right]$, where $\nabla_{T_t}\mathcal{L}_{\text{adv}}(\cdot)$ denotes the gradient of the adversarial loss with respect to the texture at step $t$.
    We update the adversarial texture $T$ using the expected gradient as: $T_{t+1} \! = \! \text{clip}\left(T_t - \eta \cdot {\mathbf{g}}_t/\|\mathbf{g}_t\|_2,\,0,\,1\right)$, where $\eta$ is the learning rate. The texture values are clipped to the valid range $[0,1]$. This iterative optimization yields an adversarial texture maintaining attack efficacy across variations in camera viewpoints.

    \noindent
    \subsubsection{\textbf{Differentiable Rendering}}
    Optimizing the adversarial texture $T$ requires the gradient of the adversarial loss $\mathcal{L}_\text{adv}$ with respect to the texture.
    However, standard robot simulators rely on non-differentiable operations, such as rasterization, making gradient computation challenging~\cite{wang2022fca, suryanto2022dta}.
    To mitigate this, we employ a hybrid rendering strategy to bypass the simulator's non-differentiable components during adversarial object optimization. 
    Specifically, the overall scene is rendered with the standard simulator, while the adversarial object $O_\text{adv}$ is rendered separately using a differentiable renderer $R$. 
    The final composed image $I_\text{adv}$ is defined as:
    $
    I_\text{adv} = (1 - M_\text{adv})\odot I_\text{sim} + M_\text{adv} \odot I_\text{diff},
    $
    where $I_\text{sim}$ and $I_\text{diff}$ denote images from the standard and differentiable renderers, respectively, $M_\text{adv}$ is a binary mask for the adversarial object, and $\odot$ denotes the Hadamard product. 
    This hybrid approach enables gradient computation with respect to the adversarial texture, enabling iterative texture optimization.
    
    \subsection{Coarse-to-Fine Attack Strategy}
    \label{sec:c2f}

    In robotic manipulation with wrist-mounted cameras, the camera-object distance varies constantly and significantly.
    While adversarial textures must remain effective across all viewing distances, simultaneous optimization for multiple distances often leads to conflicting objectives that degrade overall attack performance~\cite{cheng2024full}.
    To this end, we propose a distance-based Coarse-to-Fine (C2F) optimization strategy.
    Our approach leverages a key observation that the optimizable texture features depend on viewing distance due to changes in apparent resolution.
    At longer distances, primarily low-frequency (coarse) features remain distinguishable, while at shorter distances, high-frequency (fine) details can be effectively optimized.

    Reflecting this distance dependence, we adopt a Coarse-to-Fine (C2F) optimization strategy, optimizing sequentially from far-range to near-range scenarios. 
    In the initial \textbf{Coarse Stage}, we first optimize low-frequency components at longer viewing distances to establish robust global texture patterns. 
    Building upon this foundation, the \textbf{Fine Stage} then refines high-frequency details crucial for near-distance effectiveness.    
    By prioritizing coarse global consistency before fine-level refinements, the C2F strategy is better suited for generating adversarial textures effective throughout the robot's manipulation trajectory.

    We implement this C2F progression within the EOT framework (Sec.~\ref{sec:eot}) by scheduling the sampling of initial configurations $\tau_i \in \mathcal{T}_i$ based on their distance $r_i$.
    For each optimization stage, we sample configurations with specific camera-object distances $r_i$ according to a scheduled distribution.
    To ensure smooth transitions between stages while maintaining focus on target distance ranges, we employ a Beta distribution to control sampling probability.
    Throughout the optimization process, we progressively adjust the Beta parameters $(\alpha, \beta)$ of the Beta distribution
    to systematically shift the sampling priority from longer distances (Coarse Stage) to shorter ones (Fine Stage).

    As a result, the model first learns coarse patterns that provide a stable foundation, then refines the fine details necessary for close-range effectiveness.

\section{Experimental Result} \label{Sec:4}

    \subsection{Experimental Setup}
    \label{sec:sec4-1}

    \subsubsection{Simulation and Task} 
    We conduct experiments in the SAPIEN-based ManiSkill3 framework~\cite{Xiang_2020_SAPIEN, taomaniskill3} using a floating Panda gripper with a wrist-mounted camera. 
    The target is an end-to-end visuomotor policy $\pi_\omega$ with a ResNet18 backbone~\cite{he2016deep}, trained via PPO~\cite{schulman2017proximal} to reach a target object $O_\text{goal}$.     
    For this task, we define $O_\text{goal}$ as a `tomato\_soup\_can' and $O_\text{adv}$ as a `mustard\_bottle' from the YCB dataset~\cite{calli2015ycb}.

    \subsubsection{Adversarial Texture Optimization and C2F Staging} 
    We optimize the adversarial texture for 10k iterations using online data from 10-step rollouts across 4 environments, starting with the end-effector 25–80 cm from $O_\text{adv}$.
    For the optimization process, we set hyperparameters as follows: $\eta$=0.1, $\lambda_\text{dist}$=0.1, and $\lambda_\text{saliency}$=0.01.
    The Coarse-to-Fine (C2F) strategy proceeds in five sequential stages,
    implemented via Beta scheduling with stage-specific $(\alpha, \beta)$ as (13.48, 2.39), (23.13, 10.49), (19.88, 19.88), (10.49, 23.13), and (2.39, 13.48), progressing from coarse to fine granularity. All evaluations employ the standard SAPIEN renderer.
 
    \subsubsection{Computational Resources}
    Offline texture optimization takes approximately 5 hours on an NVIDIA RTX 4090. Each of the 10k iterations takes about 1.5 seconds and 2.5\,GB of GPU memory.
    \subsubsection{Evaluation Metrics} 
    To accurately assess attack performance, we ran 500 complete 60-step episodes for each setting, starting from task-feasible initial configurations where the baseline policy could succeed. From these episodes, we computed task-level success via ASR and T-ASR, and momentary action error via $\mathcal{E}_\text{trans}$, $\mathcal{E}_\text{rot}$. Higher ASR, T-ASR, $\mathcal{E}_\text{trans}$, and $\mathcal{E}_\text{rot}$ indicate a stronger attack.
    \begin{itemize}
        \item \textbf{Attack Success Rate (ASR):} Failure rate to reach $O_\text{goal}$.
        \item \textbf{Targeted Attack Success Rate (T-ASR):} Success rate of redirecting the robot toward $O_\text{adv}$.
        \item \textbf{Translation Error ($\mathcal{E}_\text{trans}$):} Action translation deviation between original and attacked observations.
        \item \textbf{Rotation Error ($\mathcal{E}_\text{rot}$):} Action rotation deviation between original and attacked observations.
    \end{itemize}


    \subsection{Result Analysis}
    \subsubsection*{\textbf{Effectiveness of 3D Attack}} To evaluate robustness to viewpoint changes, we compared the proposed 3D attack with a 2D patch-based attack. 
    We designed a fair 2D baseline to isolate performance differences caused purely by dimensionality. This was achieved by optimizing a thin cuboid's top surface within the 3D attack's rendering and optimization environment, which avoided traditional transformations ~\cite{byun2022improving} and kept all other conditions identical.

    As shown in TABLE~\ref{tab:3D2D}, our 3D attack consistently achieved higher ASR and T-ASR. The performance gap widened at oblique angles (large $\phi$), with the 3D attack's T-ASR being over twice as high beyond $60^{\circ}$. This is because a 2D patch's projected area shrinks and distorts from such viewpoints, whereas the 3D object maintains a larger, more stable projection.
    This demonstrates the greater robustness of 3D texture optimization to viewpoint changes, making it essential for applications like eye-in-hand robotics.

\begin{table*}[t!]
    \centering
    \caption{Attack Performance (\%) Comparison: Adversarial 3D Object (Ours) vs. 2D Patch Attack Across Viewing Angles ($\phi^{\circ}$)}
    \label{tab:3D2D}
    \setlength{\tabcolsep}{2.8pt} 
    \setlength{\aboverulesep}{0.0ex}  
    \setlength{\belowrulesep}{0.0ex}  
    \vspace{-7pt}
    \begin{NiceTabular*}{\textwidth}{c|cc cc cc cc cc cc cc cc |cc}[cell-space-limits=2.0pt]
        \Xhline{1pt}
        \Block[c]{3-1}{\raisebox{-4pt}{\textbf{\makecell{Attack \\ Type}}}} & \Block{1-18}{\textbf{Results by Viewing Angle ($\phi^{\circ}$)}} \\
        \cmidrule(l{2.0pt}r{2.0pt}){2-19}
        & \Block{1-2}{0-10} & & \Block{1-2}{10-20} & & \Block{1-2}{20-30} & & \Block{1-2}{30-40} & & \Block{1-2}{40-50} & & \Block{1-2}{50-60} & & \Block{1-2}{60-70} & & \Block{1-2}{70-90} & & \Block{1-2}{Avg} \\[-1pt]
        \cmidrule(l{2.0pt}r{2.0pt}){2-3} \cmidrule(l{2.0pt}r{2.0pt}){4-5} \cmidrule(l{2.0pt}r{2.0pt}){6-7} \cmidrule(l{2.0pt}r{2.0pt}){8-9} \cmidrule(l{2.0pt}r{2.0pt}){10-11} \cmidrule(l{2.0pt}r{2.0pt}){12-13} \cmidrule(l{2.0pt}r{2.0pt}){14-15} \cmidrule(l{2.0pt}r{2.0pt}){16-17} \cmidrule(l{2.0pt}r{2.0pt}){18-19}
        & T-ASR & ASR & T-ASR & ASR & T-ASR & ASR & T-ASR & ASR & T-ASR & ASR & T-ASR & ASR & T-ASR & ASR & T-ASR & ASR & T-ASR & ASR \\
        \hline
        2D Patch  & 73.00 & 77.00 & 61.40 & 74.40 & 54.80 & 66.80 & 46.40 & 65.00 & 31.00 & 56.60 & 27.00 & 51.80 & 15.00 & 48.80 & 10.60 & 44.20 & 39.90 & 60.58 \\
        \textbf{3D Object} & \textbf{78.00} & \textbf{79.00} & \textbf{74.00} & \textbf{80.00} & \textbf{62.00} & \textbf{77.00} & \textbf{64.00} & \textbf{74.00} & \textbf{60.00} & \textbf{70.00} & \textbf{51.00} & \textbf{62.00} & \textbf{40.00} & \textbf{55.00} & \textbf{34.20} & \textbf{56.60} & \textbf{57.90} & \textbf{69.20} \\
        \Xhline{1pt}
    \end{NiceTabular*}
\end{table*}

\begin{table*}[!t]
    \centering
    \vspace{-0pt}
    \caption{Attack Performance Comparison (T-ASR($\uparrow$), ASR($\uparrow$) in \%; $\mathcal{E}_{\text{trans}}$($\uparrow$), $\mathcal{E}_{\text{rot}}$($\uparrow$) Action Errors) Across Optimization Strategies (C2F, F2C, NS) with Ablations on Auxiliary Losses (Saliency, Targeted, Pose) at Varying Initial Robot Distances (cm).}
    \label{tab:main_tab}
    \setlength{\aboverulesep}{0.0ex}  
    \setlength{\belowrulesep}{0.0ex}  
    \vspace{-7pt}
    \begin{NiceTabular*}{\textwidth}{c | c ccc | c@{\hspace{4pt}}c c@{\hspace{4pt}}c c@{\hspace{4pt}}c c@{\hspace{4pt}}c c@{\hspace{4pt}}c | c@{\hspace{4pt}}c @{\hspace{6pt}}}[cell-space-limits=1.5pt]
    
    \Xhline{1pt} 
        & \Block[c]{4-1}{\raisebox{12pt}{\makecell{\textbf{Opt.}\\\textbf{Strategy}}}} & \Block[c]{2-3}{\textbf{Adversarial Losses}} & & & \Block[c]{1-12}{\textbf{Results by Distance (cm)}} \\[-0.2pt]
        \cmidrule(lr){6-17} \vspace*{-4pt}
        & & & & & \Block{1-2}{25--36} & & \Block{1-2}{36--47} & & \Block{1-2}{47--58} & & \Block{1-2}{58--69} & & \Block{1-2}{69--80} & & \Block{1-2}{Avg} \\[-0.5pt]
        \cmidrule(lr){3-5} \cmidrule(lr){6-7} \cmidrule(lr){8-9} \cmidrule(lr){10-11} \cmidrule(lr){12-13} \cmidrule(lr){14-15} \cmidrule(lr){16-17} 
        & & Targeted & Pose & Saliency  
        & \makecell{T-ASR \\[-1pt] $\mathcal{E}_\text{trans}$} & \makecell{ASR \\[-1pt] $\mathcal{E}_\text{rot}$} 
        & \makecell{T-ASR \\[-1pt] $\mathcal{E}_\text{trans}$} & \makecell{ASR \\[-1pt] $\mathcal{E}_\text{rot}$} 
        & \makecell{T-ASR \\[-1pt] $\mathcal{E}_\text{trans}$} & \makecell{ASR \\[-1pt] $\mathcal{E}_\text{rot}$} 
        & \makecell{T-ASR \\[-1pt] $\mathcal{E}_\text{trans}$} & \makecell{ASR \\[-1pt] $\mathcal{E}_\text{rot}$} 
        & \makecell{T-ASR \\[-1pt] $\mathcal{E}_\text{trans}$} & \makecell{ASR \\[-1pt] $\mathcal{E}_\text{rot}$} 
        & \makecell{T-ASR \\[-1pt] $\mathcal{E}_\text{trans}$} & \makecell{ASR \\[-1pt] $\mathcal{E}_\text{rot}$} \\
    \hline
    \hline
        \Block[c]{2-1}{(a)} & \Block[c]{2-1}{NS} & \Block[c]{2-1}{\checkmark} & \Block[c]{2-1}{\checkmark} & \Block[c]{2-1}{\checkmark} 
        & 56.00 & 59.60 & 69.20 & 70.40 & 73.80 & 75.20 & 75.00 & 76.80 & 94.60 & 95.20 & 73.72 & 75.44\\
        & & & & & 0.33 & 0.042 & 0.34 & 0.040 & 0.31 & 0.036 & 0.29 & 0.034 & 0.26 & 0.033 & 0.31 & 0.037 \\
    \hline
        \Block[c]{2-1}{(b)} & \Block[c]{2-1}{F2C} & \Block[c]{2-1}{\checkmark} & \Block[c]{2-1}{\checkmark} & \Block[c]{2-1}{\checkmark} 
        & 54.40 & 55.80 & 59.20 & 61.60 & 67.40 & 70.60 & 64.80 & 67.40 & 89.40 & 91.40 & 67.04 & 69.36 \\
        & & & & & 0.32 & 0.034 & 0.32 & 0.034 & 0.30 & 0.032 & 0.28 & 0.030 & 0.26 & 0.029 & 0.30 & 0.032 \\
    \hline
    \hline
        \Block[c]{2-1}{(c)} & \Block[c]{2-1}{C2F} & \Block[c]{2-1}{} & \Block[c]{2-1}{\checkmark} & \Block[c]{2-1}{\checkmark}
        & 0.20 & 15.00 & 0.00 & 8.00 & 00.00 & 10.20 & 00.00 & 8.40 & 0.00 & 34.40 & 0.04 & 15.20 \\
        & & & & & 0.05 & 0.006 & 0.04 & 0.005 & 0.03 & 0.005 & 0.03 & 0.004 & 0.02 & 0.004 & 0.04 & 0.005 \\
    \hline
        \Block[c]{2-1}{(d)} & \Block[c]{2-1}{C2F} & \Block[c]{2-1}{\checkmark} & \Block[c]{2-1}{} & \Block[c]{2-1}{\checkmark}
        & 60.60 & 63.80 & 72.20 & 73.60 & 73.80 & 75.60 & 72.60 & 73.60 & 89.80 & 91.40 & 73.80 & 75.60 \\
        & & & & & 0.21 & 0.030 & 0.22 & 0.032 & 0.24 & 0.035 & 0.24 & 0.035 & 0.26 & 0.038 & 0.23 & 0.034 \\
    \hline
        \Block[c]{2-1}{(e)} & \Block[c]{2-1}{C2F} & \Block[c]{2-1}{\checkmark} & \Block[c]{2-1}{\checkmark} & \Block[c]{2-1}{}
        & 59.20 & 62.60 & 73.80 & 74.20 & 78.60 & 79.80 & 82.00 & 82.80 & 96.80 & 97.40 & 78.08 & 79.36 \\
        & & & & & 0.49 & 0.061 & 0.48 & 0.057 & 0.43 & 0.050 & 0.37 & 0.045 & 0.33 & 0.041 & 0.42 & 0.051 \\
    \hline    
        \Block[c]{2-1}{(f)} & \Block[c]{2-1}{C2F} & \Block[c]{2-1}{\checkmark} & \Block[c]{2-1}{\checkmark} & \Block[c]{2-1}{\checkmark} 
        & \textbf{61.60} & \textbf{65.60} & \textbf{76.60} & \textbf{77.00} & \textbf{82.20} & \textbf{83.40} & \textbf{84.20} & \textbf{85.20} & \textbf{97.40} & \textbf{97.60} & \textbf{80.36} & \textbf{81.76} \\
        & & & & & \textbf{0.53} & \textbf{0.066} & \textbf{0.51} & \textbf{0.060} & \textbf{0.45} & \textbf{0.054} & \textbf{0.40} & \textbf{0.048} & \textbf{0.35} & \textbf{0.044} & \textbf{0.45} & \textbf{0.055} \\
    
    \Xhline{1pt}
    \end{NiceTabular*}
    \vspace{-13pt}
\end{table*}

    \begin{figure}[t!] 
    \centering
    \includegraphics[width=\linewidth]{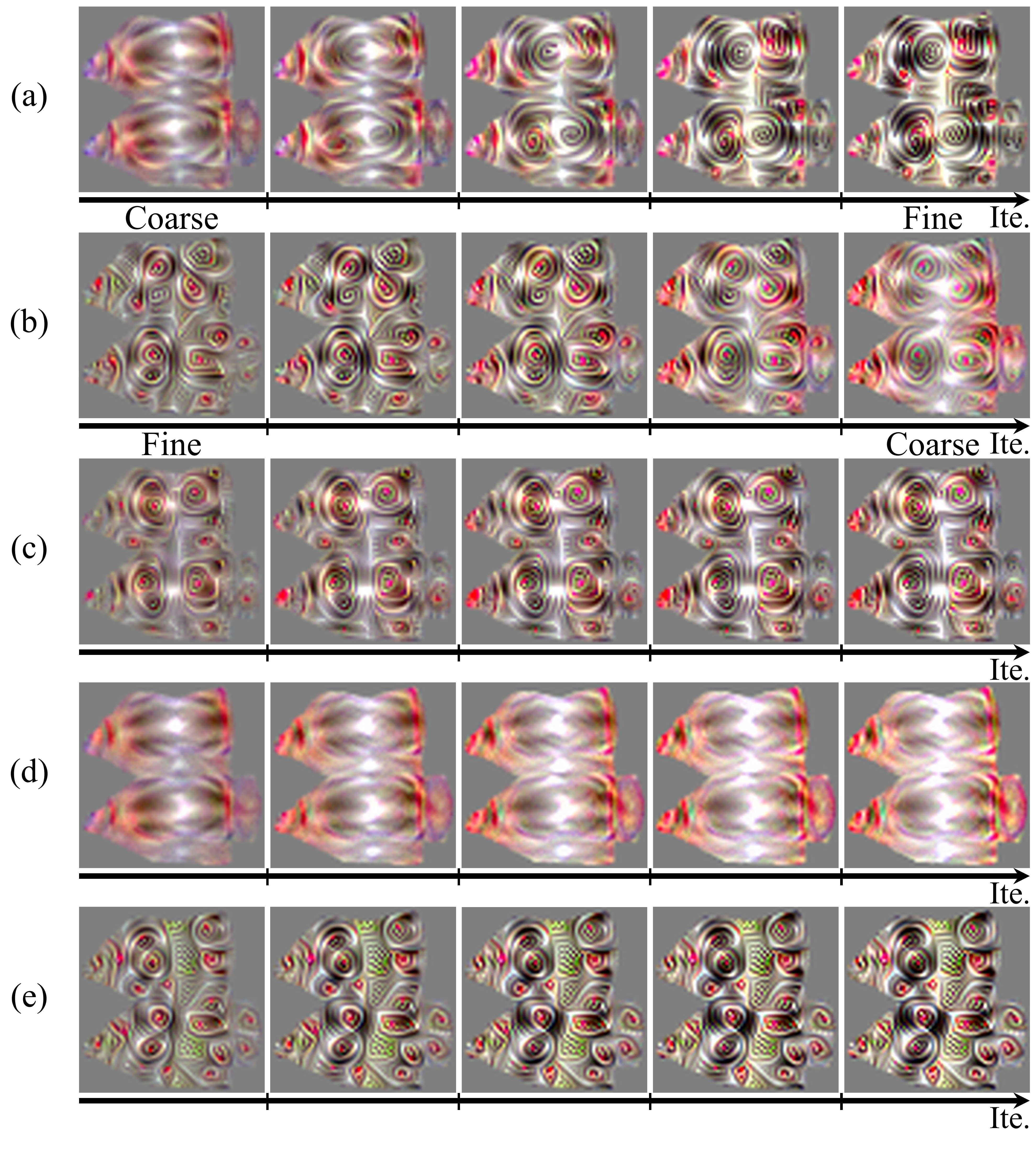} 
    \vspace{-20pt} 
    \caption{Visualization of texture update patterns under different scheduling strategies: (a) Coarse-to-Fine, (b) Fine-to-Coarse, (c) Non-staged, (d) Coarse-only, (e) Fine-only.}
    \vspace{-20pt} 
    \label{fig:texture_update} 
    \end{figure}

    \subsection{Ablation Studies}
    
    \subsubsection{\textbf{Effect of Coarse-to-Fine Optimization Strategy}}
    \label{sec:sec4-2}
    To validate our proposed Coarse-to-Fine (C2F) attack strategy, we compared its performance against several alternative optimization strategies: (1) Non-staged (uniform sampling across the distance range without staging), (2) Fine-to-Coarse (F2C) (reverse C2F order), (3) Coarse-only, and (4) Fine-only.
    Fig.~\ref{fig:texture_update} and TABLE~\ref{tab:main_tab} present their visual and quantitative comparisons, respectively. 
    The \textbf{Coarse-only} method (Fig.~\ref{fig:texture_update} (d)) produces simple textures based on low-frequency coarse features, while the \textbf{Fine-only} method (Fig.~\ref{fig:texture_update}(e)) generates intricate patterns rich in high-frequency fine details.     
    Our \textbf{C2F} method (Fig.~\ref{fig:texture_update} (a))
    creates a well-balanced texture by effectively building fine details upon a stable coarse foundation.
    Conversely, the \textbf{F2C} method (Fig.~\ref{fig:texture_update}(b)) 
    shows that initially learned fine details become blurred during the subsequent coarse stage, indicating inefficiency in the optimization path.
    The \textbf{Non-staged} approach (Fig.~\ref{fig:texture_update}(c)), which uniformly samples from all distances, results in mixed fine and coarse features throughout the optimization process.
    Quantitatively, as shown in TABLE~\ref{tab:main_tab}, the proposed C2F method not only achieved higher attack success rates (ASR and T-ASR) but also exhibited higher $\mathcal{E}_\text{trans}$ and $\mathcal{E}_\text{rot}$ values compared to the other scheduling methods.
    These results confirm that the C2F strategy, establishing global structure before refining details, is crucial for optimizing adversarial textures that maintain consistent effectiveness across the dynamic viewing conditions of wrist-mounted cameras, especially under distance variations. 
    By building a solid coarse foundation first and then adding fine details, our approach ensures reliable attack performance despite the continuous changes in viewpoint and distance inherent to wrist camera movements during manipulation tasks.

    \subsubsection{\textbf{Effect of Saliency Guidance}}
    
    \begin{figure}[!b]
    \vspace{-3pt}
    \centering
    \begin{minipage}{\linewidth} 
        \begin{tabular}{@{} m{0.02\linewidth} @{\hspace{1em}} m{\linewidth} @{}} 
        \small(a) & \includegraphics[width=0.94\linewidth, trim=0 0 6cm 0, clip]{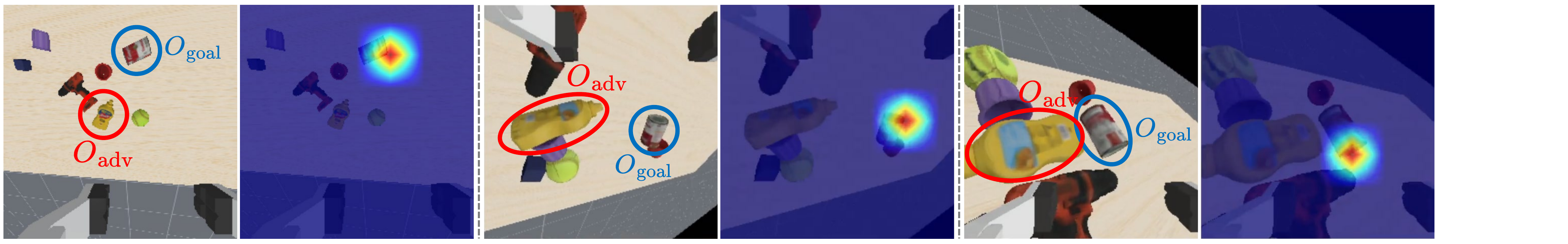}
        \end{tabular}
    \end{minipage} \\[-3pt]
    \begin{minipage}{\linewidth} 
        \begin{tabular}{@{} m{0.02\linewidth} @{\hspace{1em}} m{\linewidth} @{}} 
        \small(b) & \includegraphics[width=0.94\linewidth, trim=0 0 6cm 0, clip]{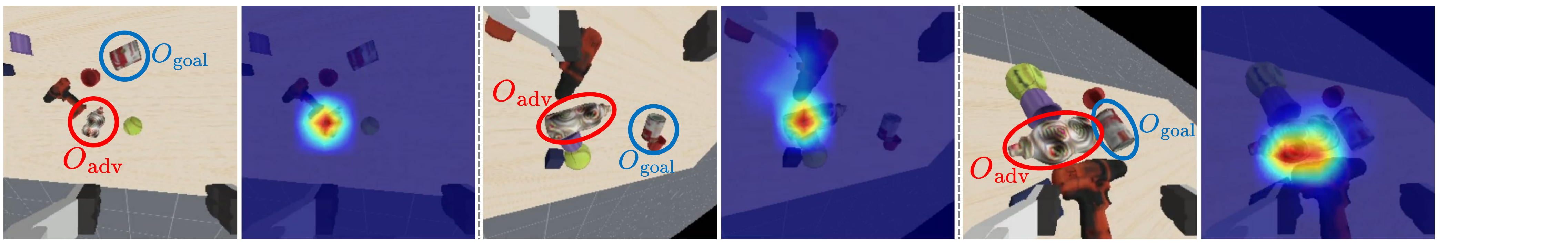}
        \end{tabular}
    \end{minipage} \\[-0pt]
    \vspace{-5pt}
    \caption{Comparison of policy saliency maps: (a) before vs. (b) after the 3D adversarial attack.} 
    \label{fig:figure5}
    \end{figure}

    This section analyzes the effect of saliency guidance, which improves optimization efficiency by focusing the attack on regions deemed important by the target policy $\pi_\omega$. 
    We evaluated its impact by comparing C2F performance with and without saliency guidance, as shown in TABLE~\ref{tab:main_tab} (d)-(f).

    Comparing TABLE~\ref{tab:main_tab} (e) and (f), we observe that incorporating saliency loss improves the attack performance across all metrics (ASR, T-ASR, $\mathcal{E}_\text{trans}$, $\mathcal{E}_\text{rot}$).
    Additionally, the results in TABLE~\ref{tab:main_tab} (d) demonstrate that using targeted saliency loss alone can achieve reasonable attack performance, confirming its individual contribution to the overall effectiveness.

    The effect of saliency guidance can be observed in the attention shift illustrated in Fig.~\ref{fig:figure5}. 
    Before the attack (Fig.~\ref{fig:figure5} (a)), the policy's attention focuses on the goal object ($O_\text{goal}$). After the attack (Fig.~\ref{fig:figure5} (b)), attention partially shifts toward the adversarial object ($O_\text{adv}$). 
    This result visually demonstrates that our attack improves performance by successfully controlling the policy's attention.

    \subsubsection{\textbf{Effect of Targeted Loss}} 
    To handle dynamic camera motion and continuously changing viewpoints, an adversarial objective must not only disrupt the policy but also keep the adversarial object $O_{\text{adv}}$ within the camera’s field of view (FOV). We evaluate our targeted loss $\mathcal{L}_{\text{adv}}$ (Sec.~\ref{sec:gradient_opt}) against an untargeted loss that simply disrupts the original goal-oriented behavior.

    The untargeted loss disrupts goal-oriented behavior by maximizing the action divergence between an adversarial image ($I_{\text{adv}}$) and a normal one ($I_{\text{sim}}$), while also reducing visual attention toward the goal object ($O_{\text{goal}}$):
    \begin{equation}
    \mathcal{L}_{\text{untargeted}} = - \|\mathbf{a} - \mathbf{a}_{\text{gt}}\|^2 + \lambda_{\text{saliency}} \cdot \frac{\sum (S \odot M_{\text{goal}})}{\sum M_{\text{goal}}}.
    \end{equation}
    
    Here, $\mathbf{a}=\pi_w(I_{\text{adv}})$ is the action on the adversarial image and $\mathbf{a}_{\text{gt}}=\pi_w(I_{\text{sim}})$ is the ground-truth action on the normal one. The loss maximizes the Mean Squared Error (MSE) between these actions to force behavioral deviation, while the saliency term penalizes attention on the goal mask ($M_{\text{goal}}$) to divert the policy's focus.

    As shown in TABLE~\ref{tab:main_tab} (f), our targeted loss $\mathcal{L}_{\text{adv}}$ yields higher ASR and T-ASR, demonstrating stronger robustness to viewpoint changes. 
    In contrast, the untargeted baseline (TABLE~\ref{tab:main_tab} (c)) shows inferior performance. 
    Due to a lack of explicit guidance toward $O_{\text{adv}}$, the attack causes only intermittent disruptions and allows the policy to recover once the adversarial object $O_{\text{adv}}$ exits the field of view. 
    These results indicate that $\mathcal{L}_{\text{adv}}$ maximizes task failure by ensuring the visibility of $O_{\text{adv}}$ through persistent guidance of the robot.

    \subsection{Generalization and Robustness Analysis}
    \label{sec:Gen_Robust}
    \begin{table}[!b]
    \vspace{-5pt}
    \centering
    \caption{Attack Generalization and Robustness Analysis (\%)}
    \vspace{-7pt}
    \setlength{\aboverulesep}{0.0ex}  
    \setlength{\belowrulesep}{0.0ex}  
    \label{tab:sim_robustness}
    \renewcommand{\arraystretch}{1.1}
    \renewcommand{\tabcolsep}{0pt} 
    \begin{tabular*}{\linewidth}{@{\hspace{.5em}} l @{\hspace{.5em}}l @{\hspace{.5em}}c@{\hspace{.5em}}c @{\hspace{.5em}}@{\extracolsep{\fill}}| l@{\hspace{.5em}} l @{\hspace{.5em}}c@{\hspace{.5em}}c @{\hspace{.5em}}}
        \toprule
        \textbf{Sec.} & \textbf{Condition} & \textbf{T-ASR} & \textbf{ASR} & 
        \textbf{Sec.} & \textbf{Condition} & \textbf{T-ASR} & \textbf{ASR} \\
        \midrule
        
         & Ours & 79.80 & 81.12 & 
        \multirow{4}{*}{\hyperref[subsec:robust]{D.3}} 
        & Light: Bright & 77.04 & 84.70 \\
        \cmidrule{1-4} 
        \multirow{2}{*}{\hyperref[subsec:gen]{D.1}}
        & Shape: Dog & 60.04 & 65.80 &
        & Light: Dim & 67.80 & 72.20 \\
        
        & Shape: Duck & 63.20 & 68.32 &
        & Add Noise & 75.57 & 80.41 \\
        \cmidrule{1-4} 
        \hyperref[subsec:transfer]{D.2}
        & Stereo Cam & 63.28 & 78.91 &
        & Bkg Varied & 77.91 & 85.64 \\

        \bottomrule
    \end{tabular*}
\end{table}

\begin{figure}[!b]
    \vspace{-12pt}
    \centering
    \subfloat[]{%
      \includegraphics[width=0.3\linewidth]{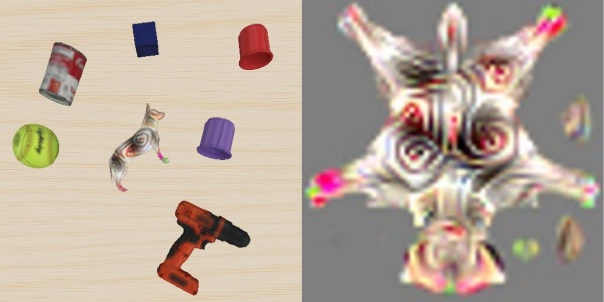}%
    }
    \hfill
    \subfloat[]{%
      \includegraphics[width=0.3\linewidth]{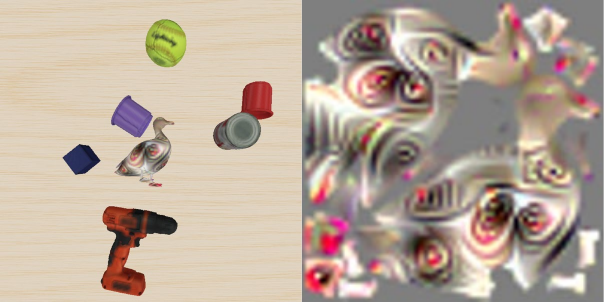}%
    }
    \hfill
    \subfloat[]{%
      \includegraphics[width=0.3\linewidth]{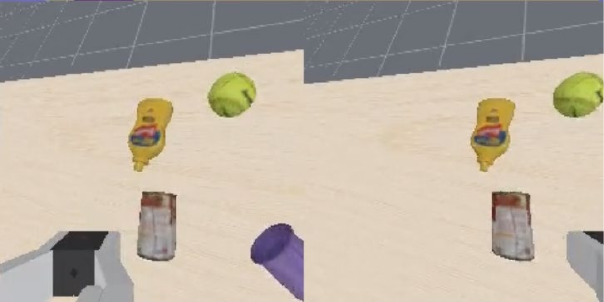}%
    }
    \vspace{-5pt}
    \caption{Visualization of Sec.~\ref{sec:Gen_Robust} configurations: (a) \hyperref[subsec:gen]{D.1}. Dog shape, (b) \hyperref[subsec:gen]{D.1}. Duck shape, (c) \hyperref[subsec:transfer]{D.2}. Stereo camera setup.}
    \label{fig:real_texture}
\end{figure}

    \subsubsection{\textbf{Generalization to Diverse Object Geometries}}
    \label{subsec:gen} 
    To verify the generalization performance of our proposed attack, we experimented with objects of various geometric structures, such as dog and duck shapes, as they are everyday objects with distinctly different and complex morphologies. This result confirms that our method is not overfitted to a specific geometry, demonstrating its ability to generate effective adversarial features for any given morphology.

    \subsubsection{\textbf{Transferability to Different Camera Configurations}}
    \label{subsec:transfer} 
    We evaluated the transferability of our single wrist-camera attack on a stereo camera setup. The attack transferred effectively, maintaining high T-ASR and ASR, which demonstrates its robustness against significant changes in camera configuration.

    \subsubsection{\textbf{Robustness to Environmental Variations}}
    \label{subsec:robust} 
    We evaluated the attack's robustness against visual environmental changes, including lighting (Light:Bright, Light:Dim), background (Bkg Varied), and sensor noise (Add Noise). The experimental results showed only minimal performance degradation, demonstrating that the proposed method operates stably even in environments with realistic variations.
    
    \noindent The results for Sec.~\ref{sec:Gen_Robust} are summarized in TABLE~\ref{tab:sim_robustness}.
    
    \noindent\textit{Note:} Detailed experimental settings and results for this Sec.~\ref{sec:Gen_Robust} are available in the supplementary video.

    \subsection{Validation in Realistic Scenarios}
    
    \subsubsection{\textbf{Transferability to Black-Box Scenario}}
    \begin{table}[!t]
    \vspace{3pt}
    \centering
    {
    \caption{Attack Transferability Across Visual Policy Architectures (Source: ResNet18$^\dagger$, (\%))}
    \label{tab:architecture_transfer_results}
    \vspace{-7pt}
    {
    \renewcommand{\arraystretch}{1.1}
    \setlength{\aboverulesep}{0.0ex}  
    \setlength{\belowrulesep}{0.0ex}  
    \begin{tabular*}{\linewidth}{@{\hspace{.5em}} @{\extracolsep{\fill}} l cc @{\hspace{1.em}} | @{\hspace{1.em}} l cc @{\hspace{.5em}}}
    \toprule 
    \textbf{Policy} & \textbf{T-ASR} & \textbf{ASR} & \textbf{Policy} & \textbf{T-ASR} & \textbf{ASR} \\
    \midrule
    ResNet18\,$^\dagger$ & 79.08 & 81.12 & Inception-v3 & 56.68 & 70.28 \\
    VGG16     & 59.68 & 70.08 & ResNet34  & 71.20 & 78.32 \\  
    \bottomrule
    \end{tabular*}%
    }
}
\vspace{-15pt}
\end{table}

    We further evaluate our method on a more practical black-box scenario where adversaries lack knowledge about the architecture of the target policy.
    To do so, we measure the ability of our attack optimized on a white-box source model using ResNet18 with PPO to fool unseen black-box target models using Inception-v3, VGG16, or ResNet34.
    As shown in TABLE~\ref{tab:architecture_transfer_results}, the attack success rates remain relatively high across these diverse architectures.
    This demonstrates that our method generalizes beyond the specific source architecture and poses a viable threat in realistic black-box scenarios.
    
    \begin{figure}[!b]
    \vspace{-15pt}
    \centering
    \subfloat[]{%
      \includegraphics[width=0.18\linewidth]{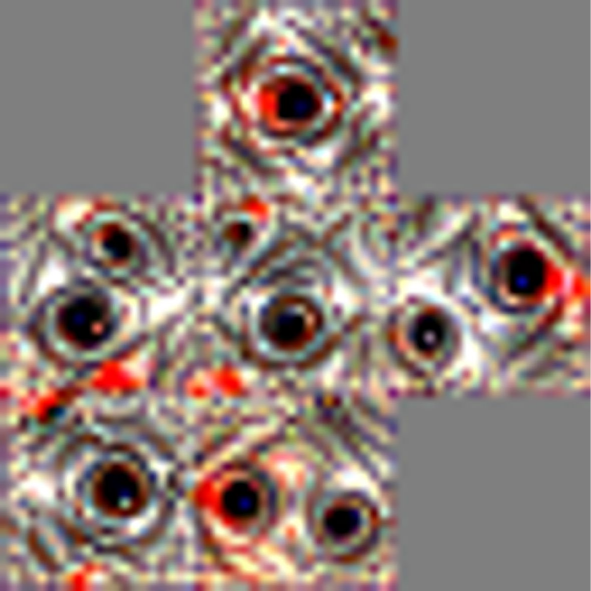}%
    }
    \hfill
    \subfloat[]{%
      \includegraphics[width=0.18\linewidth]{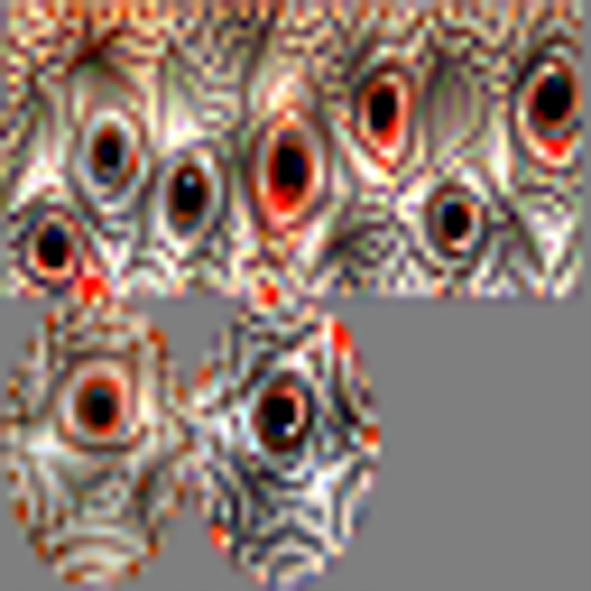}%
    }
    \hfill
    \subfloat[]{%
      \includegraphics[width=0.18\linewidth]{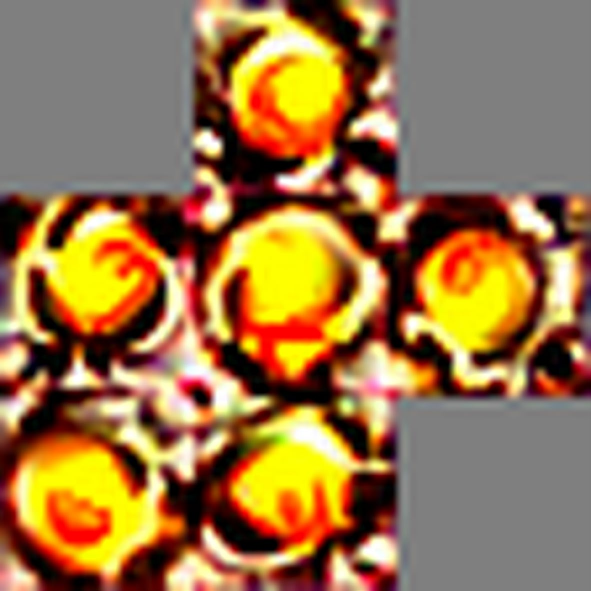}%
    }
    \hfill
    \subfloat[]{%
      \includegraphics[width=0.18\linewidth]{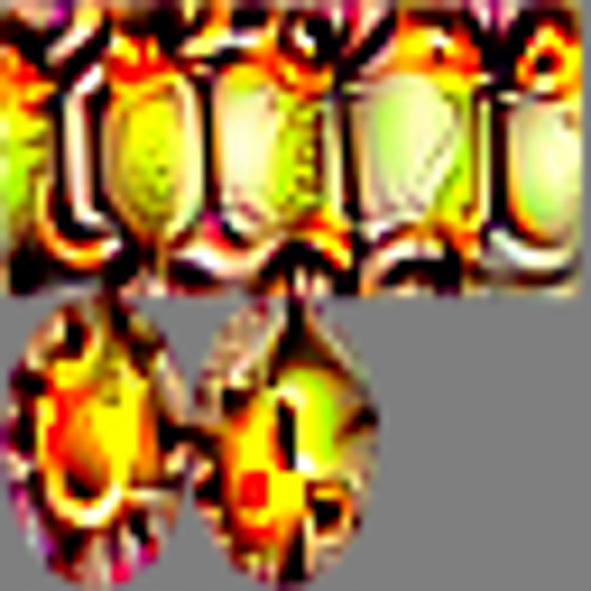}%
    }

    \vspace{-3pt}
    \caption{Visualization of adversarial textures on real-world objects: (a) Cube in EnvA; (b) Cylinder in EnvA; (c) Cube in EnvB; (d) Cylinder in EnvB.}
    \label{fig:real_texture}
    \vspace{3pt}
\end{figure}
    \begin{table}[b!]
    \centering
    \setlength{\aboverulesep}{0ex}
    \setlength{\belowrulesep}{0ex}
    \caption{Attack Sim-to-Real Transferability Evaluation (\%)}
    \vspace{-7pt}
    \label{tab:real}
    \renewcommand{\arraystretch}{1.1}
    \begin{tabularx}{\linewidth}{ c >{\raggedright\arraybackslash}X cc cc l }
        \toprule
        \multirow{2}{*}{\makecell[l]{\textbf{Env}}} & \multirow{2}{*}{\makecell[l]{\textbf{Shape} \textbf{of} \\[-1pt] \boldmath$O_{\text{adv}}$}} & \multicolumn{2}{c}{\textbf{Simulation}} & \multicolumn{2}{c}{\textbf{Real-World}} & \multirow{2}{*}{\makecell[l]{\textbf{Note}}} \\ 
        \cmidrule(lr){3-4} \cmidrule(lr){5-6}
        & & \textbf{T-ASR} & \textbf{ASR} & \textbf{T-ASR} & \textbf{ASR} \\[-1pt]
        \midrule
        A & Cube      & 71.60 & 80.60 & 60.00 & 73.33 & Fig.~\ref{fig:figure1}(b) \\
        A & Cylinder  & 76.20 & 84.60 & 73.33 & 76.67 & Fig.~\ref{fig:real}(a) \\[-0.5pt]
        \midrule
        B & Cube      & 72.80 & 83.00 & 50.00 & 60.00 & Fig.~\ref{fig:real}(b) \\
        B & Cylinder  & 75.40 & 85.40 & 53.33 & 63.33 & Fig.~\ref{fig:real}(c)\\[-0.5pt]
        \bottomrule
    \end{tabularx}
    \vspace{-5pt}
\end{table}

\begin{figure}[b!]
    \centering
    \includegraphics[width=1\linewidth, trim=0 0 0 0, clip]{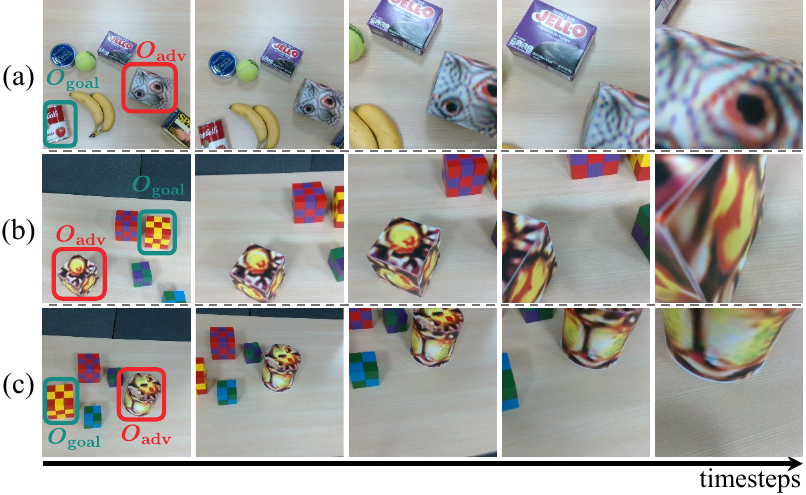}
    \vspace{-15pt}
    \caption{Policy rollout with (a) a cylinder-shaped adversarial object in EnvA, (b) a cube-shaped adversarial object in EnvB, and (c) a cylinder-shaped adversarial object in EnvB.}
    \label{fig:real}
\end{figure}

    \subsubsection{\textbf{Sim-to-Real Transferability}}
    \label{sec:sec4-6_sim2real}
    We evaluate the direct transferability of our simulation-generated adversarial objects to real-world environments. 
    To mitigate the sim-to-real gap, we incorporated lighting domain randomization during texture optimization.
    The experiments were performed with a Fetch robot and a wrist-mounted RealSense D435i camera in two settings: EnvA, where the policy aims to reach a YCB~\cite{calli2015ycb} `tomato\_soup\_can', and EnvB, where the target is a cube-shaped object (see Fig.~\ref{fig:real} (b),(c)).
    We test two 3D objects (a cube and a cylinder) for 30 trials each, covering 5 initial end-effector poses and 6 object poses.
    A reach attempt was considered successful if the gripper moved to within 10cm of the target (adversarial object $O_\text{adv}$ or goal object $O_\text{goal}$), and this target was the closest object to the gripper. 
    
    The results, shown in Fig.~\ref{fig:real} and TABLE~\ref{tab:real}, confirm that our attack is effective in the real world. 
    Despite a slight performance degradation due to sim-to-real gaps like lighting, shadows, and printing quality, the adversarial objects successfully misled the policy.

    \begin{table}[b!]
    \vspace{-7pt}
    \centering
    \caption{Attack Robustness (\%) Under Challenging Scenarios}
    \label{tab:challenge}
    \vspace{-7pt}
    \renewcommand{\arraystretch}{1.1}
    \setlength{\aboverulesep}{0.0ex}  
    \setlength{\belowrulesep}{0.0ex}  
    \begin{tabular*}{\linewidth}{
        @{\hspace{.5em}} @{\extracolsep{\fill}} 
        c @{\hspace{.5em}}
        l @{\hspace{.3em}}
        c @{\hspace{.5em}}
        c @{\hspace{.5em}} | @{\hspace{.01em}}
        c @{\hspace{.5em}}
        l @{\hspace{.3em}}
        c @{\hspace{.5em}}
        c @{\hspace{.5em}}
    }
        \toprule
        \textbf{Env} & \textbf{Scenario} & \textbf{T-ASR} & \textbf{ASR} & \textbf{Env} & \textbf{Scenario} & \textbf{T-ASR} & \textbf{ASR} \\[-2pt]
        \midrule
        A & \makecell[l]{Dynamic\\[-1pt] object (a)} & 46.67 & 60.00 & A & \makecell[l]{Occluded\\[-1pt] scene (b)} & 43.33 & 56.67 \\[-0.5pt]
        \midrule
        B & \makecell[l]{Dynamic\\[-1pt] object} & 40.00 & 50.00 & B & \makecell[l]{Occluded\\[-1pt] scene} & 36.67 & 50.00 \\[-0.5pt]
        \bottomrule
    \end{tabular*}
    \vspace{-5pt}
\end{table}

\begin{figure}[b!]
    \centering
    \includegraphics[width=1\linewidth, trim=0 0 0 0, clip]{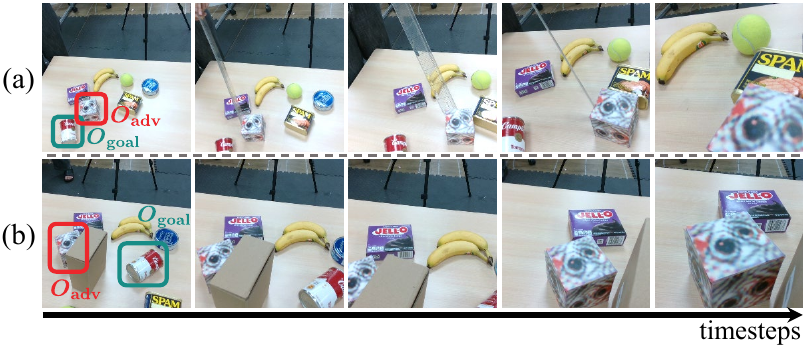}
    \vspace{-15pt}
    \caption{Policy rollout in EnvA: (a) with the dynamically moved adversarial object; (b) with the partially occluded adversarial object.}
    \label{fig:challenge}
    \vspace{7pt}
\end{figure}
 
    \begin{table}[b!]
    \vspace{-3pt}
    \centering
    \caption{Attack Robustness (\%) Under Different Lighting Variations}
    \vspace{-7pt}
    \label{tab:realworld_conditions}
    \renewcommand{\arraystretch}{1.1}
    \setlength{\aboverulesep}{0.0ex} 
    \setlength{\belowrulesep}{0.0ex} 
    
    \begin{tabular*}{\linewidth}{
        @{\hspace{.5em}} @{\extracolsep{\fill}}
        c @{\hspace{.5em}}
        l @{\hspace{.1em}}
        c @{\hspace{.5em}}
        c @{\hspace{.5em}} | @{\hspace{.01em}}
        c @{\hspace{.5em}}
        l @{\hspace{.1em}}
        c @{\hspace{.5em}}
        c @{\hspace{.5em}}
    }
        \toprule
        \textbf{Env} & \textbf{Lighting} & \textbf{T-ASR} & \textbf{ASR} & \textbf{Env} & \textbf{Lighting} & \textbf{T-ASR} & \textbf{ASR} \\[-0.5pt]
        \midrule
        A & Dim1 & 50.33 & 66.67 & A & Bright1 & 53.33 & 73.33 \\
        A & Dim2 & 53.33 & 66.67 & A & Bright2 & 56.67 & 73.33 \\
        A & Dim3 & 53.33 & 70.00 & A & Dynamic & 36.67 & 50.00 \\
        \midrule
        B & Dim1 & 46.67 & 60.00 & B & Bright1 & 50.00 & 66.67 \\
        B & Dim2 & 50.00 & 60.00 & B & Bright2 & 53.33 & 66.67 \\
        B & Dim3 & 50.00 & 63.33 & B & Dynamic & 33.33 & 50.00 \\[-0.5pt]
        \bottomrule
    \end{tabular*}
    \vspace{-5pt}
\end{table}

\begin{figure}[b!]
    \centering
    \includegraphics[width=1\linewidth, trim=0 0 0 0, clip]{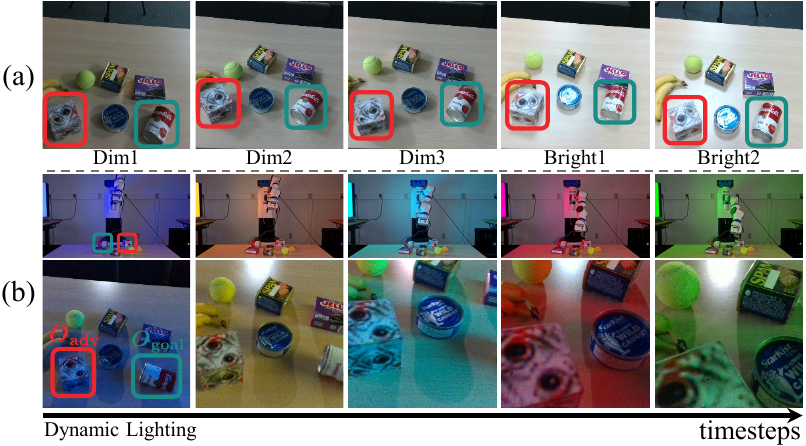}
    \vspace{-15pt}
    \caption{(a) Setups for assessing robustness to lighting in EnvA (from left: dim1, dim2, dim3, bright1, and bright2); (b) policy rollout in dynamic lighting projected via an LED monitor.}
    \label{fig:light}
\end{figure}

    \subsubsection{\textbf{Robustness in Challenging Scenarios}}
    \label{sec:sec4-7_challenging}
    To further assess the attack's resilience, we evaluate its effectiveness in more complex scenarios involving dynamically moving and partially occluded adversarial objects. For the occlusion experiments, we test scenarios where 40-70\% of the adversarial object is obscured by other objects or obstacles.
    As depicted in Fig.~\ref{fig:challenge} and TABLE~\ref{tab:challenge}, our attack maintains effectiveness even when object positions change mid-task or when objects are substantially occluded, demonstrating applicability beyond simple static settings.

    \subsubsection{\textbf{Robustness to Environmental Variations}}
    \label{sec:sec4-8_environmental}
    We assess attack performance under diverse real-world lighting conditions to evaluate robustness in practical deployment settings
    The results presented in Fig.~\ref{fig:light} and TABLE~\ref{tab:realworld_conditions} demonstrate that the attack maintains consistent effectiveness despite substantial illumination variations (dim, bright, and dynamic lighting) and cluttered or varying backgrounds.
   
    \noindent\textit{Note:} Additional experimental configurations of EnvB for Sec.~\ref{sec:sec4-7_challenging} and Sec.~\ref{sec:sec4-8_environmental} are demonstrated in the supplementary video.

\section{Conclusion} \label{sec:5}

In this paper, we propose a viewpoint-consistent 3D adversarial attack method that effectively disrupts visuomotor policies under dynamic camera viewpoints. Our method achieves this by extending 2D patches into optimized 3D objects, incorporating two key innovations: a Coarse-to-Fine (C2F) optimization strategy to ensure robustness against distance variations, and a saliency-based approach that enhances attack efficiency by targeting critical visual areas. 
Extensive experiments demonstrate not only superior performance compared to 2D patch attacks but also strong generalization across object geometries, camera setups, and black-box models.
Furthermore, our results validate the sim-to-real transferability of adversarial objects, underscoring the real-world threat posed to robotic systems.
Ultimately, our work provides both a novel attack methodology and a practical evaluation tool for strengthening the reliability of robot perception and control in safety-critical applications.

\section*{Acknowledgment}
\footnotesize
This work was supported by the Korea government (MSIT) through the Institute of Information \& Communications Technology Planning \& Evaluation (IITP) grant (RS-2025-25443318) and the National Research Foundation of Korea (NRF) grant (RS-2023-00208506).

\bibliographystyle{IEEEtran}
\bibliography{IEEEabrv,bibliography}

\end{document}